\documentclass[graybox]{svmult}

\usepackage{mathptmx}       % selects Times Roman as basic font
\usepackage{helvet}         % selects Helvetica as sans-serif font
\usepackage{courier}        % selects Courier as typewriter font
\usepackage{type1cm}        % activate if the above 3 fonts are
\usepackage{subfigure}                            % not available on your system
\usepackage{makeidx}         % allows index generation
\usepackage{graphicx}        % standard LaTeX graphics tool
\usepackage{multicol}        % used for the two-column index
\usepackage[bottom]{footmisc}% places footnotes at page bottom

\makeindex             % used for the subject index
\begin{document}

\title*{Natural Color Image Enhancement based on Modified Multiscale Retinex Algorithm and Performance Evaluation using Wavelet Energy}
\titlerunning{Natural Color Image Enhancement using Modified MSR Algorithm}% for an abbreviated version of
\author{M. C Hanumantharaju, M. Ravishankar and D. R Rameshbabu}
\institute{M. C Hanumantharaju \at Department of ISE, Dayananda Sagar College of Engineering, Shavige Malleshwara Hills, Bangalore \email{mchanumantharaju@gmail.com},
\and M. Ravishankar \at Department of ISE, Dayananda Sagar College of Engineering, Shavige Malleshwara Hills, Bangalore \email{ravishankarmcn@gmail.com}
\and D. R Rameshbabu \at Department of CSE, Dayananda Sagar College of Engineering, Shavige Malleshwara Hills, Bangalore \email{bobrammysore@gmail.com}}
\maketitle

\abstract*{This paper presents a new color image enhancement technique based on modified MultiScale Retinex(MSR) algorithm and visual quality of the enhanced images are evaluated using a new metric, namely, wavelet energy. The color image enhancement is achieved by downsampling the value component of HSV color space converted image into three scales (normal, medium and fine) following the contrast stretching operation. These downsampled value components are enhanced using the MSR algorithm. The value component is reconstructed by averaging each pixels of the lower scale image with that of the upper scale image subsequent to upsampling the lower scale image. This process replaces dark pixel by the average pixels of both the lower scale and upper scale, while retaining the bright pixels. The quality of the reconstructed images in the proposed method is found to be good and far better then the other researchers method. The performance of the proposed scheme is evaluated using new wavelet domain based assessment criterion, referred as wavelet energy. This scheme computes the energy of both original and enhanced image in wavelet domain. The number of edge details as well as wavelet energy is less in a poor quality image compared with naturally enhanced image. Experimental results presented confirms that the proposed wavelet energy based color image quality assessment technique efficiently characterizes both the local and global details of enhanced image.}

\abstract{This paper presents a new color image enhancement technique based on modified modified MultiScale Retinex (MSR) algorithm and visual quality of the enhanced images are evaluated using a new metric, namely, Wavelet Energy (WE). The color image enhancement is achieved by downsampling the value component of HSV color space converted image into three scales (normal, medium and fine) following the contrast stretching operation. These downsampled value components are enhanced using the MSR algorithm. The value component is reconstructed by averaging each pixels of the lower scale image with that of the upper scale image subsequent to upsampling the lower scale image. This process replaces dark pixel by the average pixels of both the lower scale and upper scale, while retaining the bright pixels. The quality of the reconstructed images in the proposed method is found to be good and far better then the other researchers method. The performance of the proposed scheme is evaluated using new wavelet domain based assessment criterion, referred as WE. This scheme computes the energy of both original and enhanced image in wavelet domain. The number of edge details as well as WE is less in a poor quality image compared with naturally enhanced image. Experimental results presented confirms that the proposed wavelet energy based color image quality assessment technique efficiently characterizes both the local and global details of enhanced image.}


\section{Introduction}

Image processing is a 2D-signal processing which improves the characteristics, properties and parameters of an input image in order to produce a true output picture more suitable than the input image. The important image processing operation includes enhancement, reconstruction and compression. Among these operations, image enhancement is the key step which modifies the attributes of an image to make it more appropriate for display, analysis and further processing in an image processing system. The realm of image enhancement wraps up restoration, reconstruction, filtering, segmentation, compression and transmission. Image enhancement algorithms are mainly used to pick up some important features in an image. For instance, image sharpening is done in order to bring out the details such as car license plate number, edge or line enhancement to reconstruct the objects in an aerial image and highlighting the region of interest in medical images for pathology detection of various lesions etc. These enhancement operations needs highly efficient, integrated algorithm with less number of parameters to specify. The applications such as high definition telivision, video conferencing, remote sensing etc., handles huge volumes of image data owing to increased complexity in processing. Development of image enhancement algorithm for these applications are imperative. It is of paramount importance to design an image enhancement algorithm suitable for the applications handling large image data and offer better enhancement. 

Image enhancement techniques include a filtering operation for reducing the noise present in images, contrast stretching to stretch the range of intensity values, Histogram Equalization (HE) operation to increase the contrast of an image by increasing the dynamic range of intensity values etc., The goal of the image enhancement is to extract the true image of the recorded scene. The discrepancies present in the recorded pictures described earlier are overcome in the present work by using efficient image enhancement techniques. The image enhancement algorithms mainly used in spatial domain are HE \cite{histogram}, Adaptive HE \cite{adapthist}, intensity transformations \cite{intentransform}, homomorphic filtering \cite{homomorphic} and MSR with Color Restoration (MSRCR) \cite{jobson}. Rahman et al. \cite{rahman} proposed state-of-the-art image enhancement techniques for most commonly used image enhancement techniques and validated other enhancement schemes with the MSRCR.

This paper is organized as follows: Section 2 gives a brief review of existing work. Section 3 describes the proposed modified multiscale retinex algorithm. Section 4 provides experimental results and discussions. Finally conclusion arrived at is presented in Section 5.   

\section{Existing Work}

Faming et al. \cite{faming} proposed a new pixel based variational model for remote sensing multisource image fusion using gradient features. Although multisource image fusion method adapted in this work offers integration of multiple sources, visual inspection of the reconstructed image reveals distortion in the spectral information while merging the multispectral data. Chan et al. \cite{chan} proposed fast MSR algorithm using dominant Single Scale Retinex (SSR) in weight selection. This scheme describes an approach to reduce the computational complexity in the conventional MSR algorithm. Authors claim that the quality reconstructed pictures are similar to that of conventional MSR method. Although the algorithm offers fast enhancement, subjective evaluation of the obtained results indicate the  presence of blocking artifacts, owing to poor visual quality. 

Qingyuan et al. \cite{qing} proposed an improved MSR algorithm for medical image enhancement. In this scheme, Y-component of medical image is separated into edge and non-edge area subsequent to RGB to YIQ color space conversion. The MSR technique has been used for the non-edge area in order to accomplish the medical image enhancement. However, this method provides satisfactory results for immunohistochemistry images but this approach may not provide inevitable results for other medical images. Real time modified retinex image enhancement algorithm with hardware implementation has been proposed by Hiroshi et al. \cite{hiroshi}, in order to reduce the halo artifacts. This method adaptively adjust the parameter of the cost function owing to reduced halo artifacts with improved contrast. As is seen from the experimental results presented, reconstructed images using this method are generally not satisfactory. In addition, authors have not provided the information about quality assessment of the reconstructed images.

The drawbacks of image enhancement methods specified earlier are overcome in the proposed method. The input image is first downsampled into three versions namely, normal, medium and fine scale. This downsampled images are contrast stretched to increase the picture element range and enhanced by the popular MSR algorithm. Subsequently, lower scale is upsampled to the size of next upper scale version and then combined with the next upper scale image. While combining these images, if the upsampled image has a zero pixel, then the upper scale pixel is retained otherwise, the pixel average is computed. The proposed method removes the black spots present in the Chao et al. \cite{chao} technique in an efficient way. The design developed here is much faster compared to other MSR methods since the image is downsampled into three versions. The proposed work is validated with various images of different environmental conditions. 

\section{Proposed Modified Multiscale Retinex Algorithm}

This section presents the proposed modified Multiscale Retinex based color image enhancement. The input image of resolution $256\times256$ pixels read from RGB color space is converted into Hue-Saturation-Value (HSV) color space since HSV space separates color from intensity. The value channel of HSV is scaled into three versions namely, medium scale ($64\times64$ pixels), fine scale ($128\times128$ pixels) and normal scale ($256\times256$ pixels) in order to speed up the MSR enhancement process. The hue and saturation are preserved to avoid distortion. Each of these scaled image versions may have a random pixel range. Therefore, contrast stretching operation is accomplished for each of the scaled versions of the value channel in order to translate the pixels in the display range of 0 to 255. The summary of the proposed algorithm is outlined as follows:

\begin{enumerate}
\item{Read the poor quality color image which needs image enhancement.} 
\item{Convert the image in RGB color space to HSV space to separate color from intensity.}
\item{Scale the value Component of HSV into three versions, namely, medium, fine, and normal.}
\item{Apply contrast stretching operation on each of the scaled versions to translate pixels into the display range of 0 to 255.}
\item{Apply MSR based color image enhancement of Ref. \cite{raju}}
\item{Reconstructed the value component is obtained by upsampling and combining the fine, medium and normal scale. The upsampled fine scale is combined with the medium scale by eliminating dark pixel. Similarly, the upsampled medium scale is combined with the normal scale by eliminating the dark pixel.}
\item{The reconstructed value component is combined with the preserved hue and saturation component.}
\item{Convert the image from HSV to RGB color space.}
\item{Display the enhanced Image.}
\end{enumerate}

\subsection{Multi-Scale Retinex Algorithm}

Numerous MSR based image enhancement algorithms have been reported by many researchers. However, the most popular one is the Jobson et al. \cite{jobson} MSR image enhancement algorithm since this scheme offers better image enhancement compared to other methods. The new version of the MSR algorithm developed by the Hanumantharaju et al. \cite{raju} reduces the halo artifacts of Jobson method. Therefore, in this work MSR technique proposed by Hanumantharaju et al. has been adapted in order achieve natural color image enhancement. The core part of the MSR algorithm is the design of 2D Gaussian surround function. The Gaussian surround functions are scaled in accordance with the size of the scaled value component. The size of Gaussian function employed is $64\times64$ for the medium version, $128\times128$ for the fine version and $256\times256$ for the normal version of the value channel, respectively. The general expression for the Gaussian surround function is given by Eqn. (1)

\begin{equation}
G_n(x, y) = K_n \times e^{-\frac{x^{2} + y^{2}}{2\sigma^{2}}}
\end{equation}

and $K_n$ is given by the Eqn. (2)

\begin{equation}
K_n = \frac{1}{\sum_{i=1}^{M}\sum_{j=1}^{N}{e^{-\frac{x^{2} + y^{2}}{2\sigma^{2}}}}}
\end{equation}
where x and y signify the spatial coordinates, $M \times N$ represents the image size, n is preferred as 1, 2 and 3 since the three Gaussian scales are used for each downsampled versions of the image.

Next, in order to accomplish color image enhancement, the SSR algorithm follows the MSR technique. The SSR for the value channel is given by Eqn. (3)

\begin{equation}
R_{SSRi}(x, y) = \log_2\left[V_i(x, y)\right] - \log_2\left[G_n(x, y)\otimes V_i(x, y)\right]
\end{equation}

where $R_{SSRi}(x, y)$ shows SSR output, $V_i(x, y)$ represents value channel of HSV, $G_n(x, y)$ indicated Gaussian Surround function, $\otimes$ denotes convolution operation. 

The MSR operation on a 2-D image is carried out by using Eqn. (4)

\begin{equation}
R_{MSRi}(x, y) = \sum_{n=1}^{N} W_n \times R_{SSRni}(x, y) 
\end{equation}

where $R_{MSRi}(x, y)$ shows MSR output, $W_n$ is a weighting factor which is assumed as $\frac{1}{3}$ and N indicates number of scales. 

The color image enhancement is achieved by applying the MSR algorithm for each downsampled versions subsequent to SSR operation. The new value channel is reconstructed from the individual enhanced images by combining medium, fine and normal versions of the image in an efficient way. The MSR enhanced image of medium version with resolution of $64\times64$ pixels is upsampled by two in order to match with the resolution of $128\times128$ pixels of the fine version. However, the upsampling and reconstruction operations adapted by Chao et al. \cite{chao} technique introduces zeros between alternative pixels. Although an image enhanced by this scheme is satisfactory, it has actually resulted in appearance of dots in the enhanced image and thus affects overall image quality. The present work overcomes this difficulty in a proficient way. The new fine scale version of the image is obtained as follows. The pixel of the medium scale version is retained for the zeros encountered in the upsampled medium version of the image. If there are no zeros in the upsampled medium version image than the pixel average is computed between upsampled medium version and fine version. This is illustrated by the detailed flow chart presented in Fig. 1. Finally, the composite enhanced image is reconstructed by combining new value channel with that of hue and saturation channels and converting back into RGB color space.  

\begin{figure}
\centering
\subfigure[]{
\includegraphics[height = 2in, width = 2.1 in]{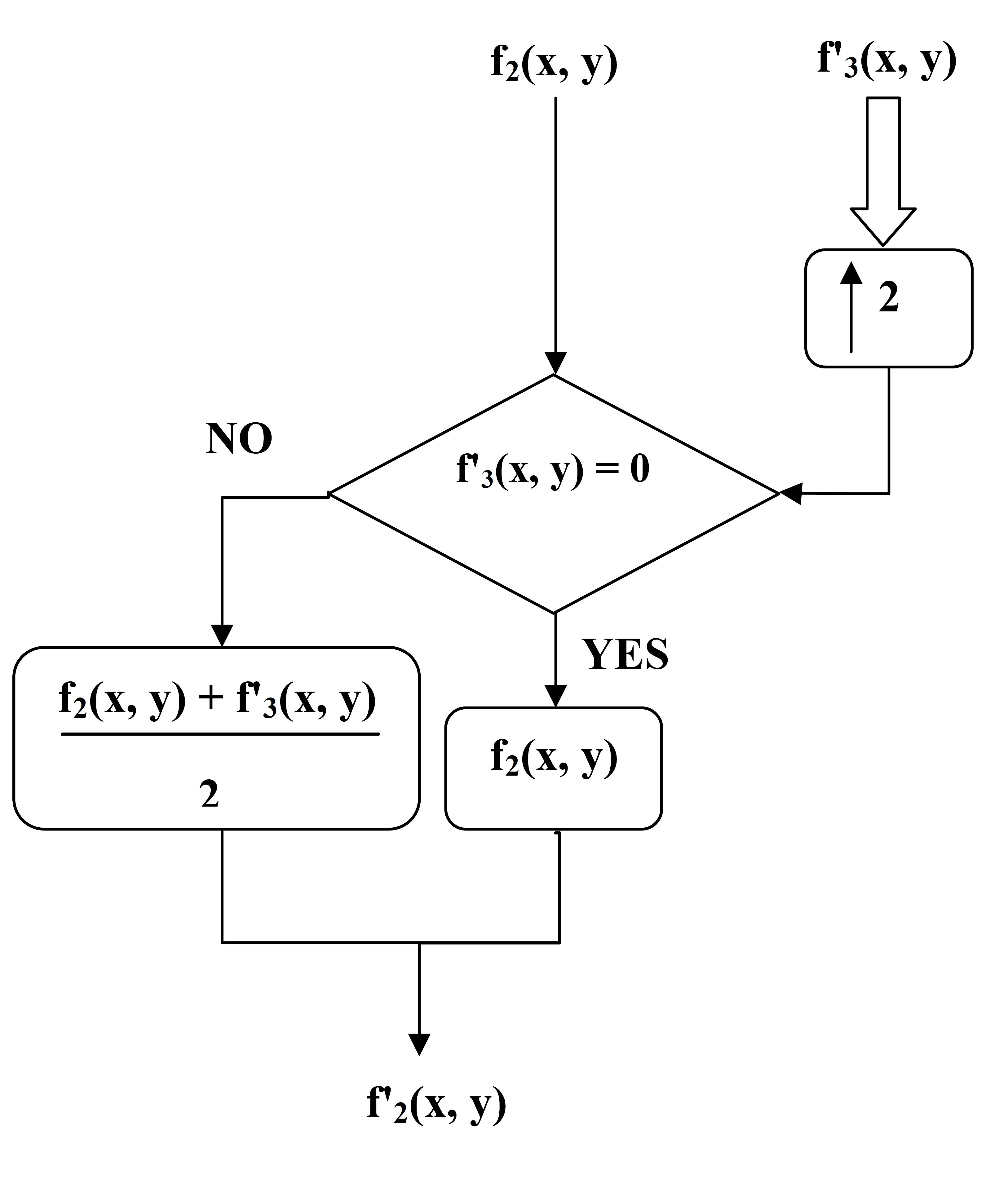}
}
\subfigure[]{
\includegraphics[height = 2in, width = 2.1 in]{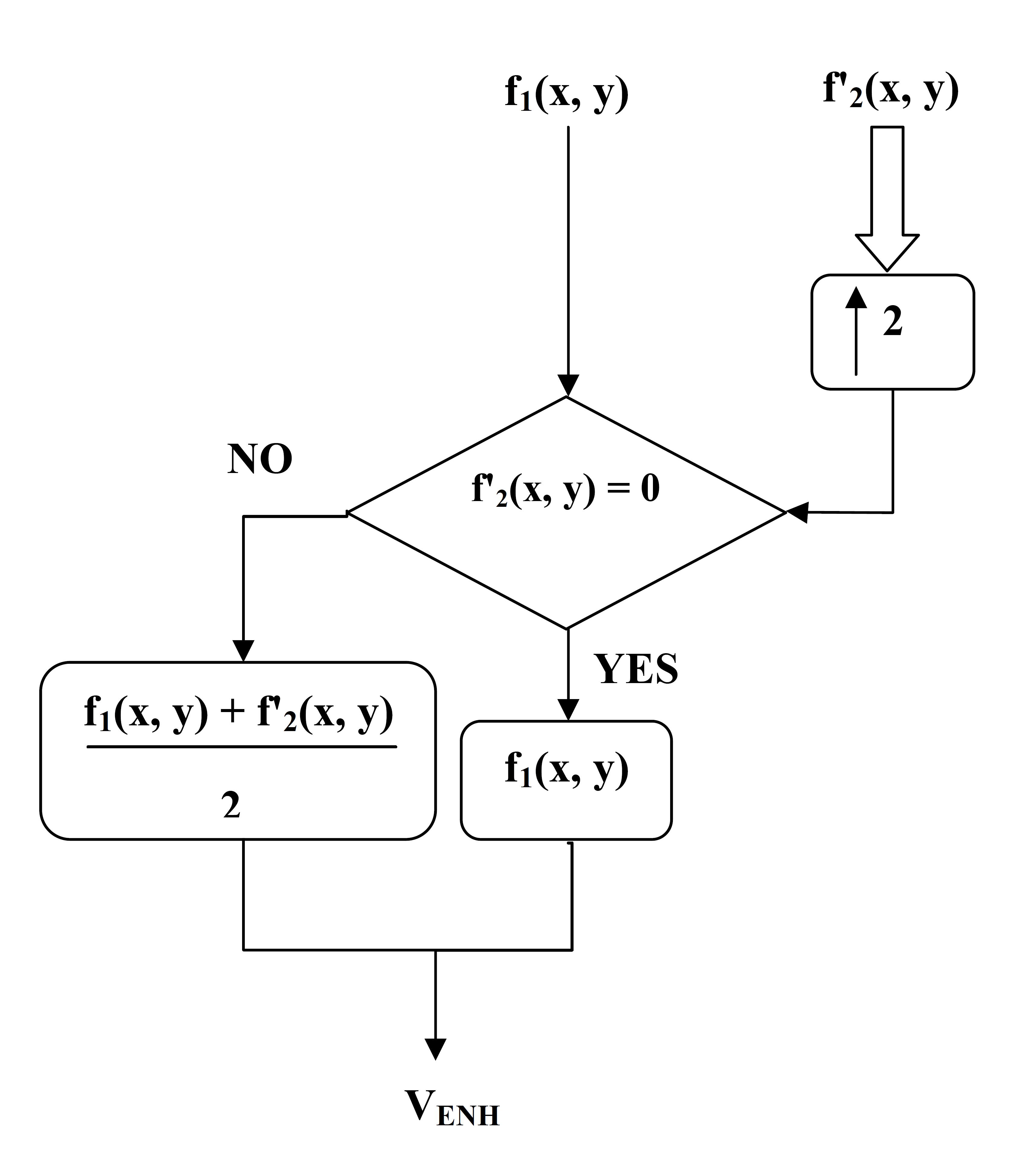}
}
\caption{Detailed Flow Sequence of the Proposed Method : (a) Flow Chart for Obtaining New Fine Scale (Resolution of $128\times128$ pixels) of Enhanced image (b) Flow Chart for Obtaining New Normal Scale (Resolution of $256\times256$ pixels) of Enhanced image}
\end{figure}

\subsection{Image Quality Assessment using Wavelet Energy}

The wavelet domain is a powerful and efficient technique for analyzing, decomposing, denoising, and compressing signals. In particular, the Discrete Wavelet Transform (DWT) breaks a signal into several time-frequency components that enables the extraction of features desirable for signal identification and recognition. The DWT and wavelet theory have been developing rapidly over the past few years. In the paper, DWT and its energy computation is exploited for visual quality assessment of an enhanced color image. The extraction of Detailed Wavelet Energy (DWE) coefficient from an image provides information about image details and the extraction of Approximate Wavelet Energy (AWE) coefficients offer the global contrast information of an image.

\subsection{Wavelet Energy}

A Continuous Wavelet Transform (CWT) maps a given function in the time domain into two dimensional function of s and t. The parameter 's' represents the scale and corresponds to frequency in Fourier transform and 't' indicates the translation of the wavelet function. The CWT is defined by Eqn. (5)

\begin{equation}
CWT(x,y) = \frac{1}{\sqrt{s}}\int S(T)\varphi\left(\frac{T-t}{s}\right)dt
\end{equation}

where S(T) is the signal and $\varphi(T)$ is the basic wavelet and $\varphi\left(\frac{T-t}{s}\right)\frac{1}{\sqrt{s}}$ is the wavelet basis function. The DWT for a signal is given by Eqn. (6)

\begin{equation}
DWT(m,n) = \frac{1}{2^{m}}\sum^{N}_{i=1}S(I,i)\phi\left[2^{-m}\left(i-n\right)\right]
\end{equation}

Wavelet energy is a method for finding wavelet energy for 1-D wavelet decomposition. The WE provides percentage of energy corresponding to the approximation and the vector containing the percentage of energy corresponding details. The WE is computed as follows

\begin{equation}
WE = \frac{1}{2^{-m/2}}\sum^{N}_{i=1}S(I,i)\phi\left[2^{-m}\left(i-n\right)\right]
\end{equation}

The WE is a Full Reference (FR) image quality assessment algorithm uses sub-band characteristics in wavelet domain. The existing metrics are analyzed and limitations are investigated. Image quality evaluation using WE computation uses a linear combination of high frequency coefficients after a Daubechies wavelet transform. The probability density function of the enhanced image has relatively higher energy in wavelet domain compared to other transforms. Therefore, the wavelet energy metric is an effective and efficient metric to evaluate the quality of the enhanced image. The approximate wavelet energy coefficients provide the information on the global image enhancement and detailed WE coefficients provide statistics on the image details. 

\section{Experimental Results and Comparative Study}

The algorithm presented in the earlier section was coded using Matlab Version 8.0. The experiment was conducted by considering poor quality images of different environmental conditions downloaded from various databases. The first column of Fig. 2 presents the original image. The same images enhanced using NASA's MSR scheme of Ref. \cite{jobson} is shown in second column of Fig. 2. The improved MSR scheme of Ref. \cite{shen} is presented in third column of Fig. 2. Finally, reconstructed images using proposed modified MSR based scheme is shown in last column of Fig. 2. 

In order to show the efficiency of the proposed method in more detail, the algorithm is tested with other test images. The first column of Fig. 3 shows the original image. The second column of Fig. 3 shows the same image enhanced using MSRCR of Ref. {jobson}. The third column presents the image enhanced using improved MSRCR of Ref. {shen}. The last column of Fig. 3 shows the image enhanced using proposed modified MSR method. 

\begin{figure}
\centering
\subfigure{
\includegraphics[height = 0.9 in, width = 1 in]{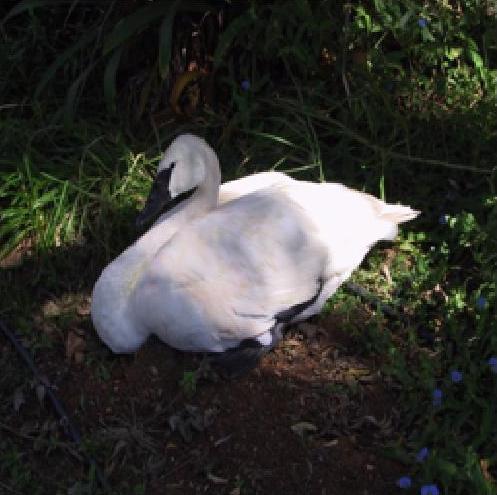}
}
\subfigure{
\includegraphics[height = 0.9 in, width = 1 in]{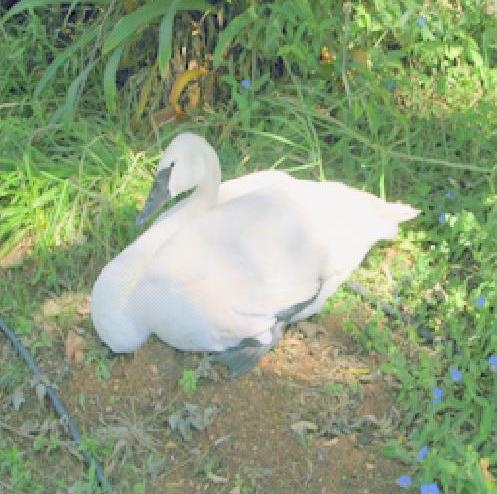}
}
\subfigure{
\includegraphics[height = 0.9 in, width = 1 in]{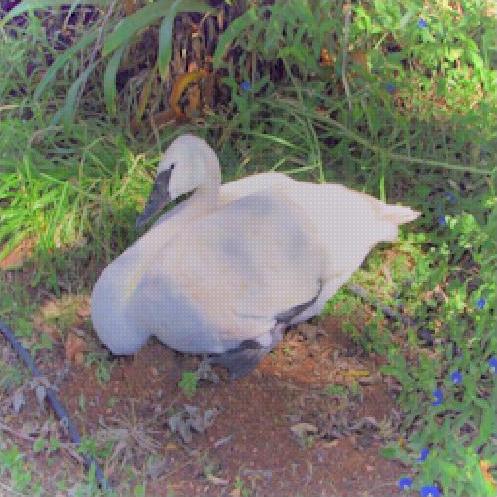}
}
\subfigure{
\includegraphics[height = 0.9 in, width = 1 in]{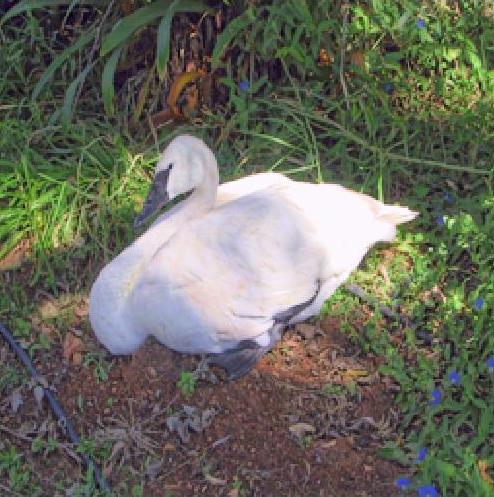}
}
\subfigure{
\includegraphics[height = 0.9 in, width = 1 in]{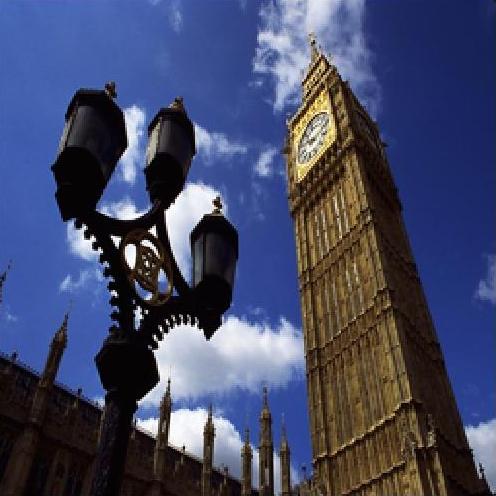}
}
\subfigure{
\includegraphics[height = 0.9 in, width = 1 in]{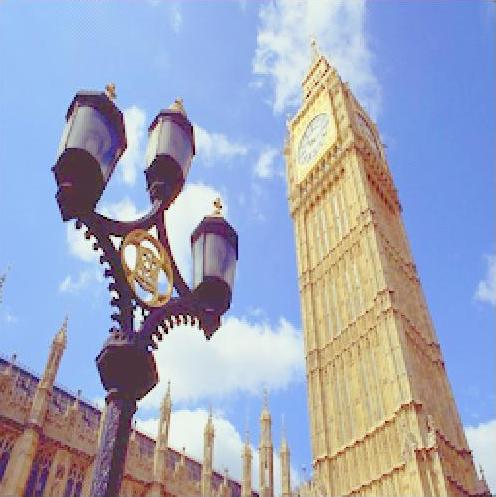}
}
\subfigure{
\includegraphics[height = 0.9 in, width = 1 in]{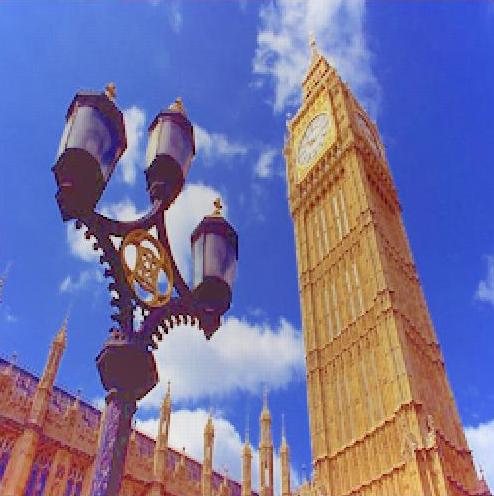}
}
\subfigure{
\includegraphics[height = 0.9 in, width = 1 in]{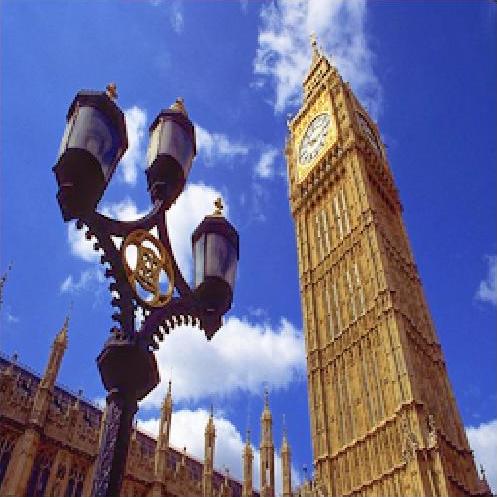}
}
\subfigure{
\includegraphics[height = 0.9 in, width = 1 in]{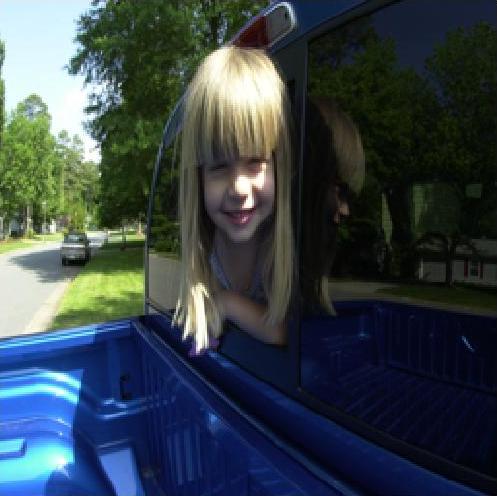}
}
\subfigure{
\includegraphics[height = 0.9 in, width = 1 in]{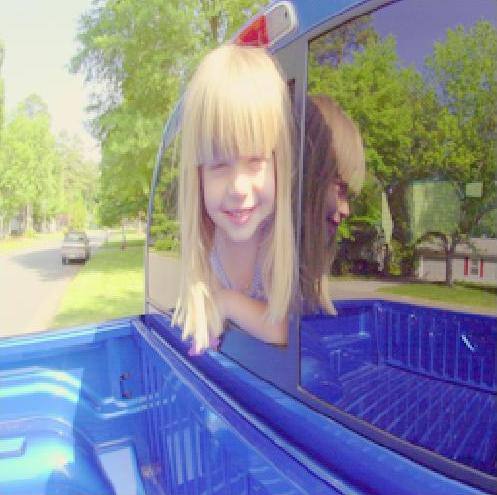}
}
\subfigure{
\includegraphics[height = 0.9 in, width = 1 in]{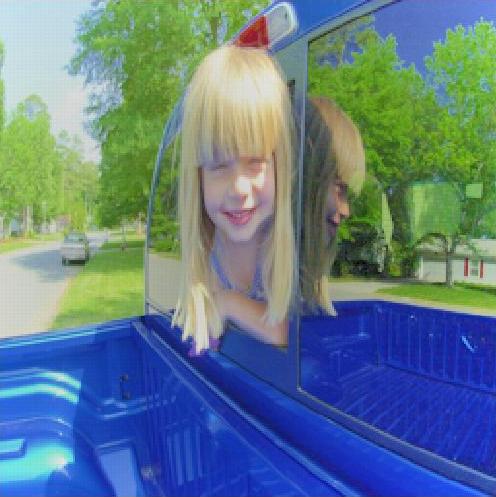}
}
\subfigure{
\includegraphics[height = 0.9 in, width = 1 in]{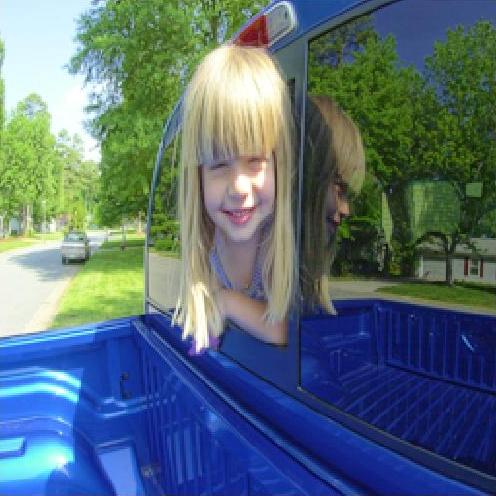}
}
\caption{\textbf{First Column:} Original Image of Resolution $256\times256$ pixels, \textbf{Second Column:} Image Enhanced using Multiscale Retinex with Color Restoration (MSRCR) of Ref.  \cite{jobson}, \textbf{Third Column:} Improved MSRCR of Ref. \cite{shen}, \textbf{Fourth Column:} Proposed Method}
\end{figure}

\begin{figure}
\centering
\subfigure{
\includegraphics[height = 0.9 in, width = 1 in]{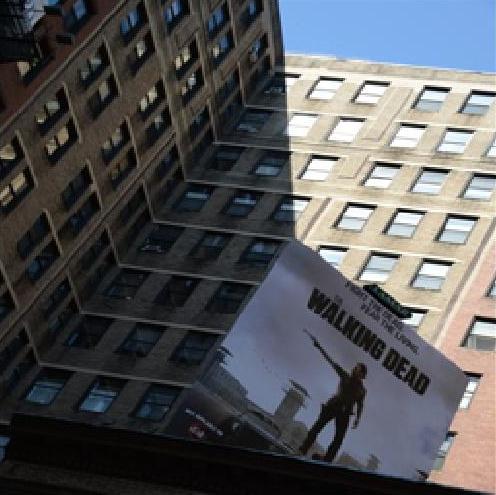}
}
\subfigure{
\includegraphics[height = 0.9 in, width = 1 in]{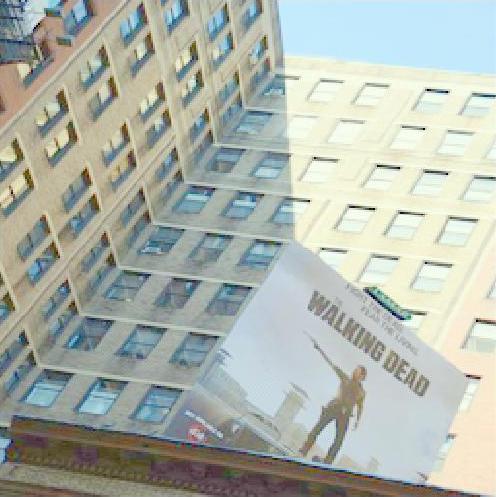}
}
\subfigure{
\includegraphics[height = 0.9 in, width = 1 in]{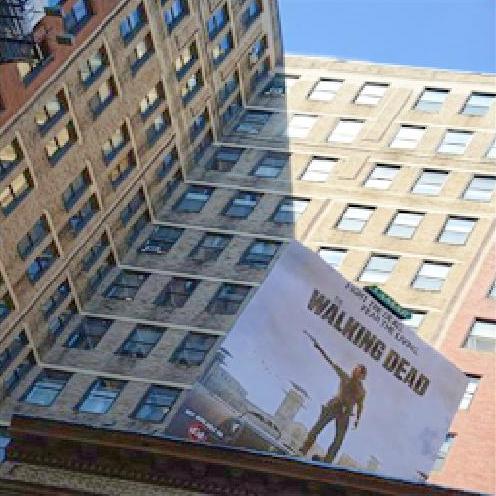}
}
\subfigure{
\includegraphics[height = 0.9 in, width = 1 in]{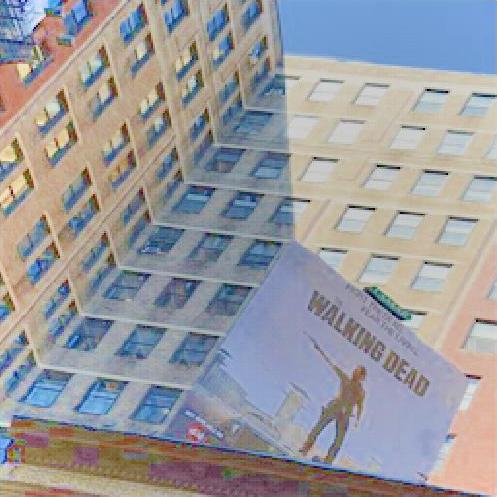}
}
\subfigure{
\includegraphics[height = 0.9 in, width = 1 in]{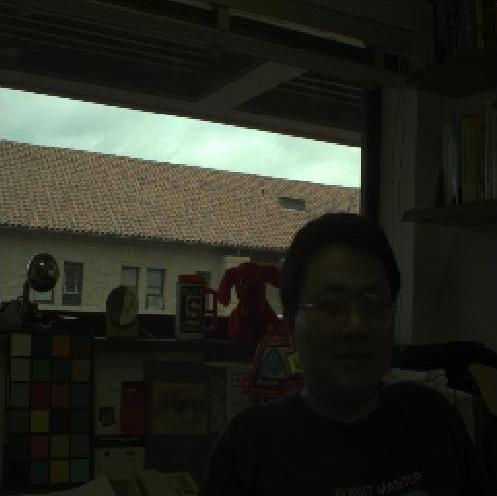}
}
\subfigure{
\includegraphics[height = 0.9 in, width = 1 in]{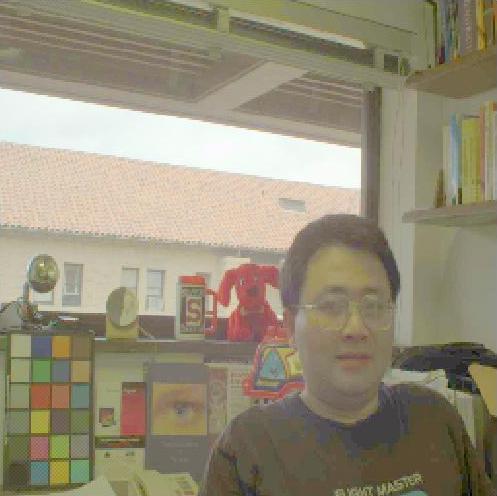}
}
\subfigure{
\includegraphics[height = 0.9 in, width = 1 in]{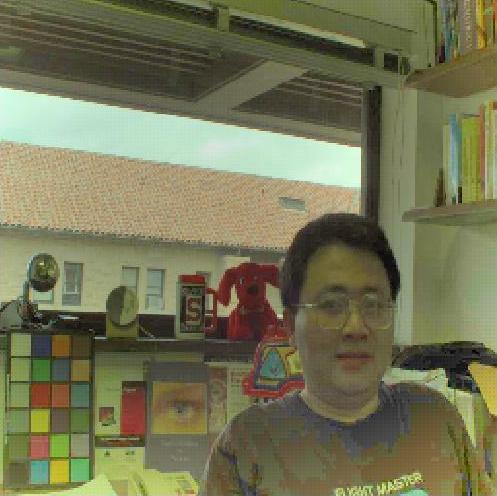}
}
\subfigure{
\includegraphics[height = 0.9 in, width = 1 in]{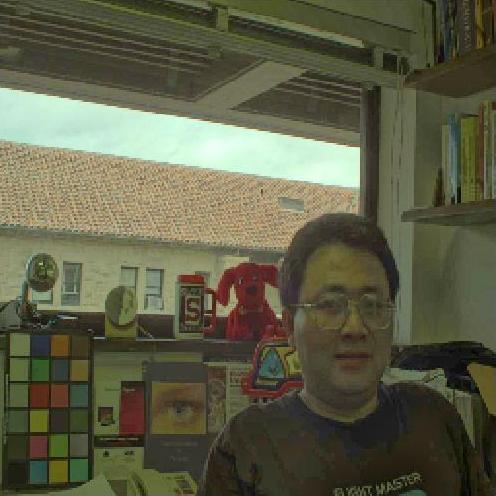}
}
\subfigure{
\includegraphics[height = 0.9 in, width = 1 in]{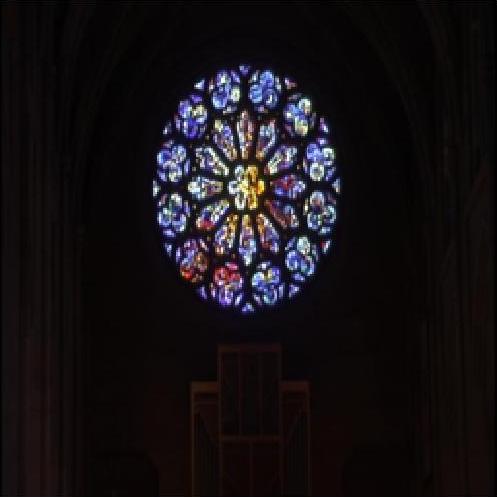}
}
\subfigure{
\includegraphics[height = 0.9 in, width = 1 in]{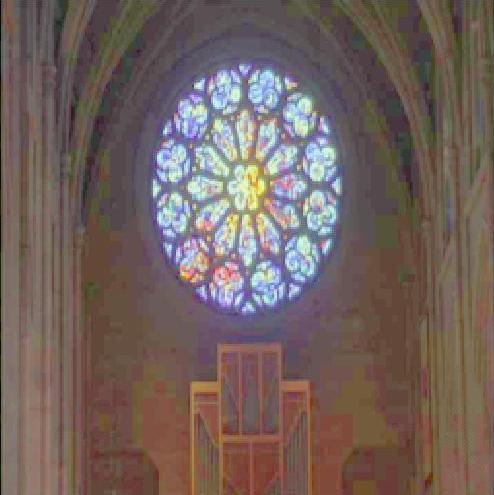}
}
\subfigure{
\includegraphics[height = 0.9 in, width = 1 in]{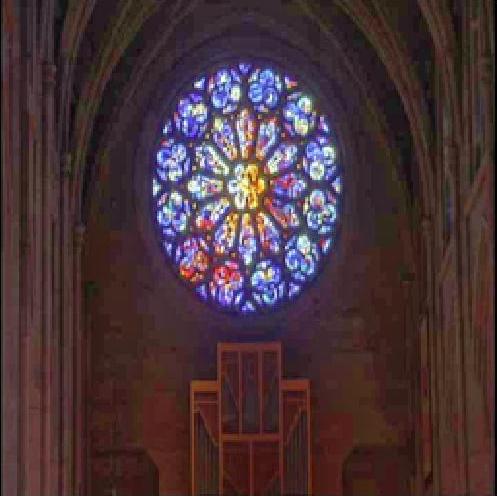}
}
\subfigure{
\includegraphics[height = 0.9 in, width = 1 in]{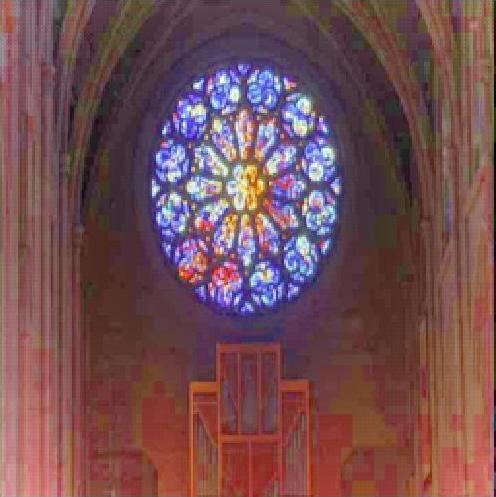}
}
\caption{\textbf{First Column:} Original Image of Resolution $256\times256$ pixels, \textbf{Second Column:} Image Enhanced using Multiscale Retinex with Color Restoration (MSRCR) of Ref.  \cite{jobson}, \textbf{Third Column:} Improved MSRCR of Ref. \cite{shen}, \textbf{Fourth Column:} Proposed Method}
\end{figure}

The image enhancement achieved using the proposed modified MSR algorithm is validated using the WE metric described earlier. Although, numerous metrics for color image quality assessment were reported by various researchers, measurement do not correlate well with image perception. Therefore, WE based IQA has been exploited in this work in order to verify the performance of the image enhancement algorithm. Table 1 shows the image quality assessment based on AWE and DWE. It may noted that the Average WE calculated for enhanced image of the proposed method is in close proximity to that of the original image. However, the image enhanced using MSRCR of Ref. \cite{jobson} and Improved MSRCR of Ref. \cite{shen} are large compared to proposed method. Similarly, it may be analyzed from the DWE presented that the proposed modified MSR algorithm has DWE close to that of original image. However, DWE calculated for other enhancement methods are smaller than the original image. This shows that the proposed modified MSR based image enhancement method offers a natural enhancement which may be perceived by the Human Visual System (HVS).   

\begin{table}
\centering
\caption{Approximate (A) and Detailed (D) WE Metric for Color Image Quality Assessment, \textbf{Note:} The highlighted values in the table shows better enhancement method based on the proposed wavelet energy metric. In the evaluation, first preference is given for details followed by overall enhancement.}
\begin{tabular}{|l|c|c|c|c|c|c|c|c|}
\hline 
\multicolumn{1}{|c|}{\textbf{}} &
\multicolumn{2}{|c|}{\textbf{Org. Image}} &
\multicolumn{2}{|c|}{\textbf{MSRCR}} &
\multicolumn{2}{|c|}{\textbf{Improved MSRCR}} &
\multicolumn{2}{|c|}{\textbf{Proposed Method}} \\%\multicolumn{2}{|c|}{\textbf{Detl. Coefficient}}  \\ 
\hline
\textbf{Test Images} &\textbf{AWE}& \textbf{DWE}& \textbf{AWE}& \textbf{DWE}& \textbf{AWE}& \textbf{DWE}& \textbf{AWE}& \textbf{DWE} \\
\hline 
\textbf{Swan}&99.57 &0.423  &99.84  &0.154 &99.76  &0.2375 &99.62	&0.374    \\ 
\hline 
\textbf{Port}&99.60 &0.394  &99.87  &0.126 &99.83  &0.163 &99.68	&0.317    \\ 
\hline
\textbf{Girl} &99.75 &0.244  &99.91  &0.087 &99.87  &0.127 &99.82	&0.175    \\ 
\hline
\end{tabular}
\label{table1}
\end{table}

\section{Conclusion}

A new color enhancement approach based on modified multiscale retinex algorithm as well as the wavelet energy metric for image quality assessment has been proposed. HSV color space is exploited since this domain separates color from intensity. The value component is downsampled into three versions, namely, normal, medium and fine. The contrast stretching operation is performed on the each downsampled value components. Subsequently, MSR algorithm used for value component enhancement. As is evident from the experimental results presented that the proposed modified MSR based color image enhancement offers natural quality images. It is found that the enhanced images are more vivid, brilliant, and correlates to the HVS. Research is in progress to implement the proposed algorithm on reconfigurable hardware device such as Field Programmable Gate Arrays (FPGAs). 

%%%%%%%%%%%%%%%%%%%%%%%% referenc.tex %%%%%%%%%%%%%%%%%%%%%%%%%%%%%%
% sample references
% %
% Use this file as a template for your own input.
%
%%%%%%%%%%%%%%%%%%%%%%%% Springer-Verlag %%%%%%%%%%%%%%%%%%%%%%%%%%
%
% BibTeX users please use
% \bibliographystyle{}
% \bibliography{}
%

\end{document}